\documentclass[sigplan,10pt, authorversion,nonacm, screen]{acmart}

\usepackage{hyperref}
\usepackage{url}

\usepackage{capt-of} 
\usepackage{caption} 
\usepackage{subcaption}
\usepackage{todonotes}
\usepackage{epsfig,nicefrac,verbatim,tabularx,xspace}
\usepackage[]{siunitx}
\usepackage{algorithm}
\usepackage[noend]{algpseudocode}

\usepackage[capitalise]{cleveref}

\usepackage{booktabs}
\usepackage{multirow}
\usepackage[normalem]{ulem}

\usepackage{flafter}

\usepackage{enumitem}

\AtBeginDocument{%
  \providecommand\BibTeX{{%
    \normalfont B\kern-0.5em{\scshape i\kern-0.25em b}\kern-0.8em\TeX}}}

\begin{document}

\title{ \textit{}\\Can Fair Federated Learning reduce the need for Personalisation?}


\author{Alex Iacob, Pedro P. B. Gusm\~{a}o, Nicholas D. Lane}
\affiliation{%
  \institution{University of Cambridge}
}




\begin{abstract}
  Federated Learning (FL) enables training ML models on edge clients without sharing data. However, the federated model's performance on local data varies, disincentivising the participation of clients who benefit little from FL. Fair FL reduces accuracy disparity by focusing on clients with higher losses while personalisation locally fine-tunes the model. Personalisation provides a participation incentive when an FL model underperforms \emph{relative} to one trained locally. For situations where the federated model provides a lower accuracy than a model trained entirely locally by a client, personalisation improves the accuracy of the pre-trained federated weights to be similar to or exceed those of the local client model. This paper evaluates two Fair FL (FFL) algorithms as starting points for personalisation. Our results show that FFL provides no benefit to relative performance in a language task and may double the number of underperforming clients for an image task. Instead, we propose Personalisation-aware Federated Learning (PaFL) as a paradigm that pre-emptively uses personalisation losses during training. Our technique shows a $50\%$ reduction in the number of underperforming clients for the language task while lowering the number of underperforming clients in the image task instead of doubling it. Thus, evidence indicates that it may allow a broader set of devices to benefit from FL and represents a promising avenue for future experimentation and theoretical analysis.

\end{abstract}

\maketitle

\rhead{}
\lhead{}

\chead{PREPRINT: Accepted at The 3rd Workshop on
  Machine Learning and Systems (EuroMLSys 20223)}
\fancypagestyle{firstpagestyle}{%
  \chead{PREPRINT: Accepted at The 3rd Workshop on
    Machine Learning and Systems (EuroMLSys 20223)}
}

\section{Introduction}
Edge devices provide computational power and data for Machine Learning tasks; however, minimising communication costs while using them can be challenging. Federated Learning (FL) was introduced by \citet{mcmahan2017communication} to enable model training on client devices without sharing data. However, the FL model accuracy may be underwhelming on clients with unusual data and even worse than a local model, reducing the incentive for participation.

The existing body of research on balancing global and local performance has proposed several approaches. Two Fair Federated Learning (FFL) techniques, q-Fair Federated Learning (q-FFL) and Tilted Empirical Risk Minimization (TERM), proposed by \citet{li2019fair} and \citet{TERM} respectively, aim to improve the accuracy of the worst-performing clients by prioritising those with large losses during FL. Alternatively, \citet{yu2020salvaging} and \citet{Mansour2020ThreeAF} recommend using personalisation (local adaptation) methods such as Freezebase (FB), Elastic Weight Consolidation (EWC), and Knowledge Distillation (KD) for fine-tuning. In this work, \emph{relative accuracy} refers to the difference in local client test set accuracy between a federated and local model.

While the sets of potential use cases for fairness and personalisation are not identical---e.g., personalisation would be inappropriate for clients with few samples---FFL could construct a fairer relative accuracy distribution as a starting point. For FFL to reduce the need for personalisation, it would have to lower the number of underperforming clients or improve their average relative accuracy. However, in our experiments, FFL had a neutral or negative effect on the relative accuracy distribution. \emph{Our contribution is threefold:}

\begin{enumerate}
    \item We construct an initial empirical evaluation of the relative accuracy distribution of models trained with FFL on the Reddit, CIFAR-10, and FEMNIST datasets for next-word prediction and image recognition tasks. During our evaluation, we show that FFL does not significantly reduce the number of underperforming clients or improve the relative accuracy distribution on Reddit and brings little benefit over FedAvg and personalisation. We also show it doubles the number of underperforming clients for FEMNIST.
    \item We investigate potential synergies between FFL and personalisation by adapting fair federated models. Results show that the adapted models do not significantly outperform those initially trained with FedAvg in relative accuracy or the number of underperforming clients.
    \item We propose Personalization-aware Federated Learning (PaFL) as a paradigm that uses local adaptation techniques during FL training to pre-empt personalisation. Results on the language task show a significant reduction in underperforming clients over FFL when applying KD  without any downsides to subsequent personalisation. Moreover, PaFL can avoid the increase in underperforming clients observed for image recognition on FEMNIST when using EWC or KD for our tested hyperparameters. However, given that our results are based entirely on simulation, future work consisting of additional experimentation and theoretical analysis is needed to determine the exact relationship between the training loss, e.g., KD or EWC, fairness, and local accuracy after personalisation.
\end{enumerate}

\section{Background and Related Work}\label{back:FLobj}

\citet{li2020federated} formulate the FL objective function as seen in \cref{eq:flObjective}
\begin{equation} \label{eq:flObjective}
    \underset{w}{\min} f(w) = \sum_{k=1}^m p_k F_k(w) \ ,
\end{equation}
where \(f\) is the federated loss, $m$ is the client count, $w$ is the model, and \(F_k\) is the loss of client \(k\) weighted by \(p_k\). For a total number of samples \(n\), \(p_k\) is defined as the proportion of samples on the client \(\tfrac{n_k}{n}\). The Federated Averaging (FedAvg) algorithm introduced by \citet{mcmahan2017communication} trains locally on clients and for each round $t$ sums the parameters of each model \(G_k^t\) from client $k$ weighted by \(p_k\) with the previous model \(G^t\) using learning rate \( \eta \), as seen in \cref{eq:FedAvg}
\begin{equation} \label{eq:FedAvg}
    G^{t+1} = G^t + \eta \left( \sum_{k=1}^m p_k G_k^t \ \right) \ .
\end{equation}

\paragraph{Data and Hardware heterogeneity} Data generation, network speed and computation naturally vary across devices due to hardware, location, time, and behaviour. These factors make the data distribution not Idendepentend and Identically Distributed (IID), leading to feature label or quantity skew as reported by \citet[sec.3.1]{kairouz2019advances}. Non-IID data can impact accuracy  \citep{zhao2018federated, non-iid-quagmire} and convergence \citep{li2019convergence} while different hardware results in stragglers and unreliability.

\subsection{Fair Federated Learning}
\citet{li2019fair} propose Fair FL (FFL), which defines a ``fairer'' FL model as one that achieves a lower variance in its accuracy distribution over local client test sets while keeping average accuracy similar. They propose a version of FFL, q-FFL, to emphasise underperforming clients during federated training as seen in \cref{eq:FFLobjective}
\begin{equation} \label{eq:FFLobjective}
    \underset{w}{\min} f(w) = \sum_{k=1}^m \cfrac{p_k}{q+1}\,F_k^{q+1}(w) \ ,
\end{equation}
where $q$ controls the degree of fairness. A $q=0$ corresponds to FedAvg, while larger values prioritise higher losses to improve accuracy on clients for which the federated model underperforms. \citet{TERM} develop Tilted Empirical Risk Minimization (TERM), shown in \cref{eq:TERMobjective}, which behaves similarly to q-FFL. While the two objectives show comparable improvements in the evaluations of \citet{TERM}, their interactions with personalisation are unknown.
\begin{equation} \label{eq:TERMobjective}
    \underset{w}{\min} f(w) = \cfrac{1}{t} \log(\sum_{k=1}^m p_k e^{tF_k(w)}) \ ,
\end{equation}
The original publications of \citet{li2019fair,TERM} used a weighted sampling of devices based on their number of examples, followed by uniform averaging. However, the server must know the dataset size of all clients a priori, which is potentially unfeasible. Thus, we use uniform sampling and weighted averaging in this work.

The most relevant recent FFL work is Ditto, published by \citet{DITTO}, which constructs personalised models while encouraging fairness for the global model. Ditto keeps a persistent local model in sync with the federated one by minimising the $L_2$ distance to the federated model, similar to the personalisation techniques discussed below. While \citet{DITTO} show this local regularisation to be superior to TERM in promoting fairness, it requires more resources than personalisation. Maintaining a persistent local model incurs training costs on every single round, in addition to the increased storage demands during training, without the benefit of a highly-trained federated model to provide pre-trained weights. As such, we have chosen to opt against maintaining persistent local models; however, we recommend them as a promising avenue for future work.
\subsection{Local Adaptation}

The analysis of \citet{yu2020salvaging} established that the federated model performs worse on heterogeneous clients, as previously noted by \citet{li2019fair, kairouz2019advances}, and it may offer inferior performance to local ones. They propose several techniques to address this.

\paragraph{Elastic-weight Consolidation (EWC)}\label{back:MTL}

The task of the global model is to maintain performance on all previous clients while training locally. \Cref{eq:MTL} frames it as Multi-task Learning (MTL) problem using the Elastic Weight Consolidation technique (EWC) introduced by \citet{kirkpatrick2017overcoming} to avoid Catastrophic
forgetting \citep{goodFellowForgetting}
\begin{equation}\label{eq:MTL}
    l(C,x) = L(C,x) + \sum_i \tfrac{\lambda}{2} M[i] (C[i] - G[i])^2 \ ,
\end{equation}
where L is the client loss, \(\lambda\) determines the weighting between the two tasks and $M$ is the Fisher information matrix.
\paragraph{Fine-tuning (FT) and Freezebase (FB)}\label{back:Freezebase}
When a client receives a global model after the FL process, it can apply Fine-tuning \citep{wangFineTuning,federated_Personalization,Mansour2020ThreeAF} to retrain the model on its data. Furthermore, to avoid potential Catastrophic forgetting, \citet{yu2020salvaging} also opt to apply Freezebase (FB) as a variant of FT which retrains only the top layer.

\paragraph{Knowledge Distillation (KD)}\label{back:KD}

As an alternative to EWC and FT, Knowledge Distillation \citep{hinton2015distilling} uses the global model as a teacher for a client model. For the pure logit outputs of the federated model \(G(x)\) and client model \(C(x)\), the client minimises the loss in \cref{eq:kdLoss}
\begin{equation}\label{eq:kdLoss}
    l(C,x) =  \alpha T^2 L(C,x) + (1-\alpha) K_L (\sigma(G(x) \, / \, T), \sigma(C(x) \, / \, T)) \ ,
\end{equation}
where \(L\) is the client loss, \(K_L\) is the Kullback-Leibler divergence \citep{KLloss}, \(\sigma \) is the softmax, \(\alpha \) is the weighting and \(T \) is the temperature.

\paragraph{FedProx}
One relevant loss function not considered by \citet{yu2020salvaging} is the constraint on the Euclidean distance of model parameters employed by FedProx~\citep{li2018federated} to limit model divergence. The loss function shown in \cref{eq:FedProx} is formulated similarly to EWC in \cref{eq:MTL}.
\begin{equation}\label{eq:FedProx}
    l(C,x) = L(C,x) + \sum_i \tfrac{\lambda}{2}  (C[i] - G[i])^2 \ ,
\end{equation}
Rather than using it to create personalised models, the original work of \citet{li2018federated} employs this loss function between the model parameters at the start of the round and the parameters being trained on the client, which serves as the inspiration for our proposal.

\section{Personalisation-aware Federated Learning}

As an alternative to FFL for reducing personalisation costs, we consider modifying local client training in a manner which pre-empts a later adaptation phase by using loss functions meant for personalisation. The procedure is roughly analogous to quantisation-aware training \citep{QuantAwareTraining}. This work uses \emph{Personalisation-aware Federated Learning (PaFL)} to refer to such a paradigm. While Federated Learning and local adaptation have historically been regarded as separate, the FedProx algorithm developed by \citet{li2018federated} may be considered prototypical to PaFL as it injects the $L_2$ norm of the model weight differences into the local loss.

The FedProx algorithm aims to mitigate model divergence caused by data heterogeneity in a manner that may improve the accuracy of the federated model on average across clients. However, it only considers the contribution of the model parameters based on their magnitude rather than their importance to the output of the federated model. While this choice is well-justified by \citet{li2018federated}, we consider some modifications while maintaining the principle when adapting it to pre-empt personalisation.

Personalisation-aware Federated Learning modifies FedProx to allow the loss function and the afferent weight to vary across rounds. Beyond potentially improved convergence, such a process may benefit final locally-trained models by providing continuity in the local objective between FL training and the final adaptation stage if the same loss function is used. Furthermore, loss-based weighted averaging as used in q-FFL (\cref{eq:FFLobjective}) and TERM (\cref{eq:TERMobjective}) has no means of reconciling differences between models required by clients with equally high losses and highly divergent data partitions. By contrast, PaFL allows clients for whom the global model performance is underwhelming to diverge in a manner which maintains accuracy on the whole federated distribution on which the model was trained.

Formally, PaFL can be defined as a type of Federated Learning where each client has a loss function obeying the structure in \cref{eq:PaFL}:
\begin{equation}\label{eq:PaFL}
    l(C,x, t) = \mu(t) \, L(C,x) + ( 1 - \mu(t)) \, D(t)(C, G, x) \ ,
\end{equation}
where \(t\) is the current round, \(L(C,x) \) is the training loss and \(D(t)\) returns a personalisation loss function for the current round---potentially dependent on the data point \(x\). The weight of each term is set per round through the weighting function \( \mu(t)\). PaFL can be naturally extended to incorporate local adaptation if deemed beneficial by allowing clients to keep their model received after the final training round. However, for the rest of this work, PaFL shall refer only to the federated training phase. Local adaptation happens after federated training is complete to allow comparison against combinations of FL algorithms and personalisation methods.
\section{Experimental Design}
Following the lead of \citet{yu2020salvaging} and \citet{mcmahan2017communication}, we train models using FedAvg, q-FedAvg, TERM, or PaFL for next-word prediction on a version of the Reddit~\citep{caldas2019expanding} dataset with \(80\,000\) participants each having $150-500$ posts treated as sperate sentences. We also train on CIFAR-10 partitioned into $100$ participants and the naturally heterogeneous Federated Extended MNIST (FEMNIST) \citep{caldas2019expanding} for image recognition. During local adaptation, we follow the parameters recommended by \citet{yu2020salvaging}. For EWC, we use a weighting of $\lambda=5000$; for KD, we use a temperature $T=6$ and weighting $\alpha=0.95$. Our hyperparameter choices attempt to replicate those of \citet{yu2020salvaging} whenever possible.

\paragraph{Reddit} Reddit contains diverse sentences from its forum users, which makes it a valuable resource for Federated Learning, as users' total word counts and vocabulary size vary across several orders of magnitude with a skewed distribution. We train a standard LSTM for next-word prediction using  $2$ layers, $200$ hidden units and 10 million parameters. To construct tokens, we employ the dictionary of the $50,000$ most frequent words compiled by \citet{yu2020salvaging}; all other words are replaced with placeholders.  The first \(90\% \) of a user's posts, chronologically, is used as a training set, with the final \(10\% \) reserved for local testing. A separate centralised test set is maintained for evaluating global task performance during the FL training process with \(\approx 5\% \) of it used to track convergence. In contrast, the full test set is used for the final evaluation. Federated models train for \(1\,000\) rounds using \(20\) clients per round. On the client side, models train for \(2\) internal epochs with a batch size of \(20\) using SGD with a learning rate of $40$. For adaptation, we use a learning rate of \(1\) and batch size of \(20\) for \(100\) epochs of retraining.

\paragraph{FEMNIST} Federated Extended MNIST is an image dataset comprised of $62$ characters written in a $28x28$ format. It is naturally divided into clients based on the author of a character, with each client having $226$ samples on average. We use a similar experimental setup to \citet{caldas2019expanding} with a simple two-layer CNN. Rather than subsampling \(5\%\) of the data from all clients as \citet{caldas2019expanding} do, we keep \(350\) clients with more than \(10\) samples out of the total \(3\,597\). We use \(70\%\) of a client's data for training, $10\%$ for local testing and add the remaining \(20\%\) to the federated test set. For the FL process, we use an aggregation learning rate of \(\eta = 1.0\) with \(10\) clients per round for \(500\) rounds. During training, we use SGD for $2$ internal epochs with a learning rate of \(0.1\) and a batch size of \(32\) for each client, while during adaptation, we lower the learning rate to \(0.01\).

\paragraph{CIFAR-10} CIFAR-10 is an image dataset composed of $60,000$ images of $10$ objects in a $32x32$ format. Since CIFAR-10 is not a naturally federated dataset as it is not split into clients, a Dirichlet distribution (\(\alpha = 0.9\)) is used to simulate a Non-IID partitioning similarly to \citet{hsu2019} and \citet{yu2020salvaging}. A ResNet-18 \citep{ResNet} model is trained over \(1
,000\) rounds with $10$ clients per round. Clients are trained using a batch size of $32$ with \(2\) internal epochs and a learning rate of \(0.1\). The test accuracy is computed by multiplying a client's per-class accuracy on the CIFAR-10 test set with its proportion of the local device data. For adaptation, we use a learning rate of \(10^{-3}\) and batch size of \(32\) for \(200\) epochs. Training uses SGD with momentum $0.9$ and weight decay $5 \times 10^{-4}$,

\subsection{Experiments}\label{sec:experiments}
We train models for FedAvg, q-FedAvg, TERM and PaFL. Specifically, FedAvg (i.e., q-FedAvg with $q=0$) is trained using the abovementioned standard parameters and serves as the baseline for all datasets and models. For q-FedAvg, we test $q \in \{0,0.01,0.1,0.5,1,5\}$ for Reddit and show evaluation results for the relevant values of $q \in \{0,0.1,5\}$ which produce sufficiently distinguished results. Similarly, we test $q \in \{0,0.1,1,5,10,15\}$ for FEMNIST and CIFAR-10, and we report the evaluation results for $q \in \{0,10,15\}$ and $q \in \{0,5,15\}$ respectively as to showcase the overall trend that increases in fairness create. For TERM on Reddit we use $t \in \{0.1,5\}$ while on FEMNIST we do not tune the value of $t$ for TERM and instead reuse the $t=1$ value chosen by \citet{DITTO}. For PaFL we choose a simple proof-of-concept training sequence where we apply KD or EWC with constant weightings and parameters after the model has approached convergence at the halfway point of training---denoted $H_{EWC}$ and $H_{KD}$. We use the same parameters and weightings for the losses after the halfway point as in local adaptation.

\paragraph{Centralised evaluation}\label{method:fedPerf}
The first experiment uses the held-out centralised test set of each dataset defined above to test federated models. It is also used to choose which models should be tested locally or adapted given our hardware constraints from \cref{hardware}.

\paragraph{Local Accuracy Evaluation}\label{method:absolute_local_performance}

The second experiment evaluates federated models on each client's local test set and reports average accuracy and variance for the client population. We also investigate accuracy and variance for the best and worst $10\%$ of clients in terms of test accuracy. Finally, we report the average accuracy of locally trained models using the same parameters as in the local adaptation phase without the personalisation loss.

\paragraph{Local Adaptation and Relative Accuracy Evaluation}\label{method:relative_local_performance}

The primary experimental setup entails comparing the accuracy of federated or adapted models with purely local models on client data; the difference between the two is referred to as relative accuracy. The effectiveness of federated models is determined by two key factors: the number of clients with positive relative accuracy and the average relative accuracy. If a synergistic relationship exists between FFL or PaFL and local adaptation methods, models trained using these techniques would significantly improve average relative accuracy or have fewer underperforming clients after adaptation.

\subsection{Hardware Limitations}\label{hardware}
Each node of the cluster that the experiments were run on holds four Nvidia A100 GPUs. Given the quotas and service levels of the cluster, the number of clients on which the federated model could be tested locally for the language task was limited to \(\sim 65\,500\) to avoid incurring costs beyond the allocated university funds. Similarly, the number that could be adapted was limited to ($\sim\!18500$). Therefore, all charts and tables comparing local model or adaptation performance use data from the client set common to all results.

\section{Results}

\begin{figure}[]
    \centering
    \includegraphics[width=\columnwidth]{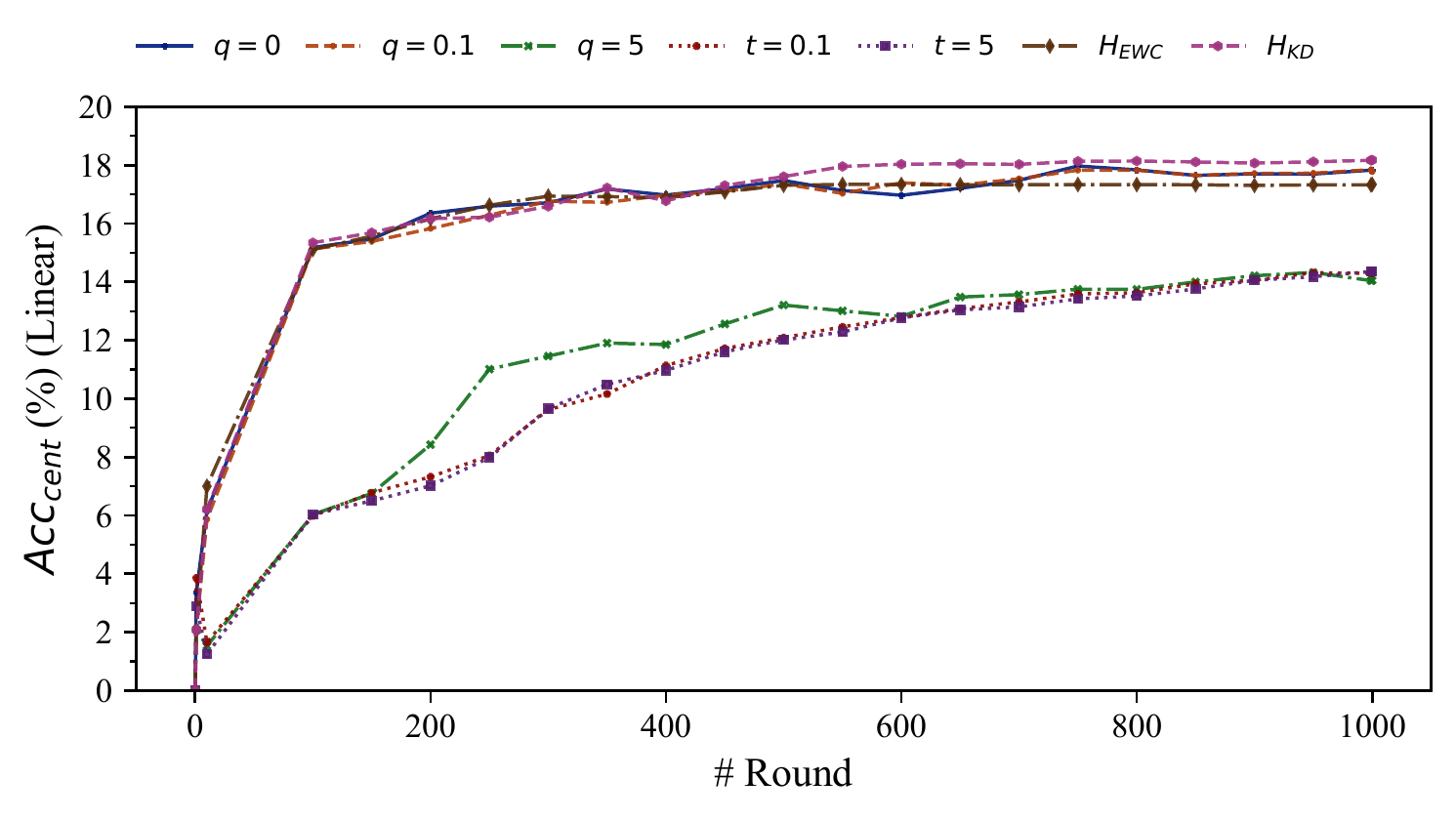}
    \caption{Language task test-set accuracy, TERM harms performance for our tested values while q-FFL only does so significantly for $q \geq 1.0$. Crucially, both $H_{EWC}$ and $H_{KD}$ approach the FedAvg baseline with $H_{KD}$ exceeding it.}
    \label{fig1:globalAGGacc:FairPlot:a}
\end{figure}

\begin{figure}[]
    \centering

    \includegraphics[width=\columnwidth]{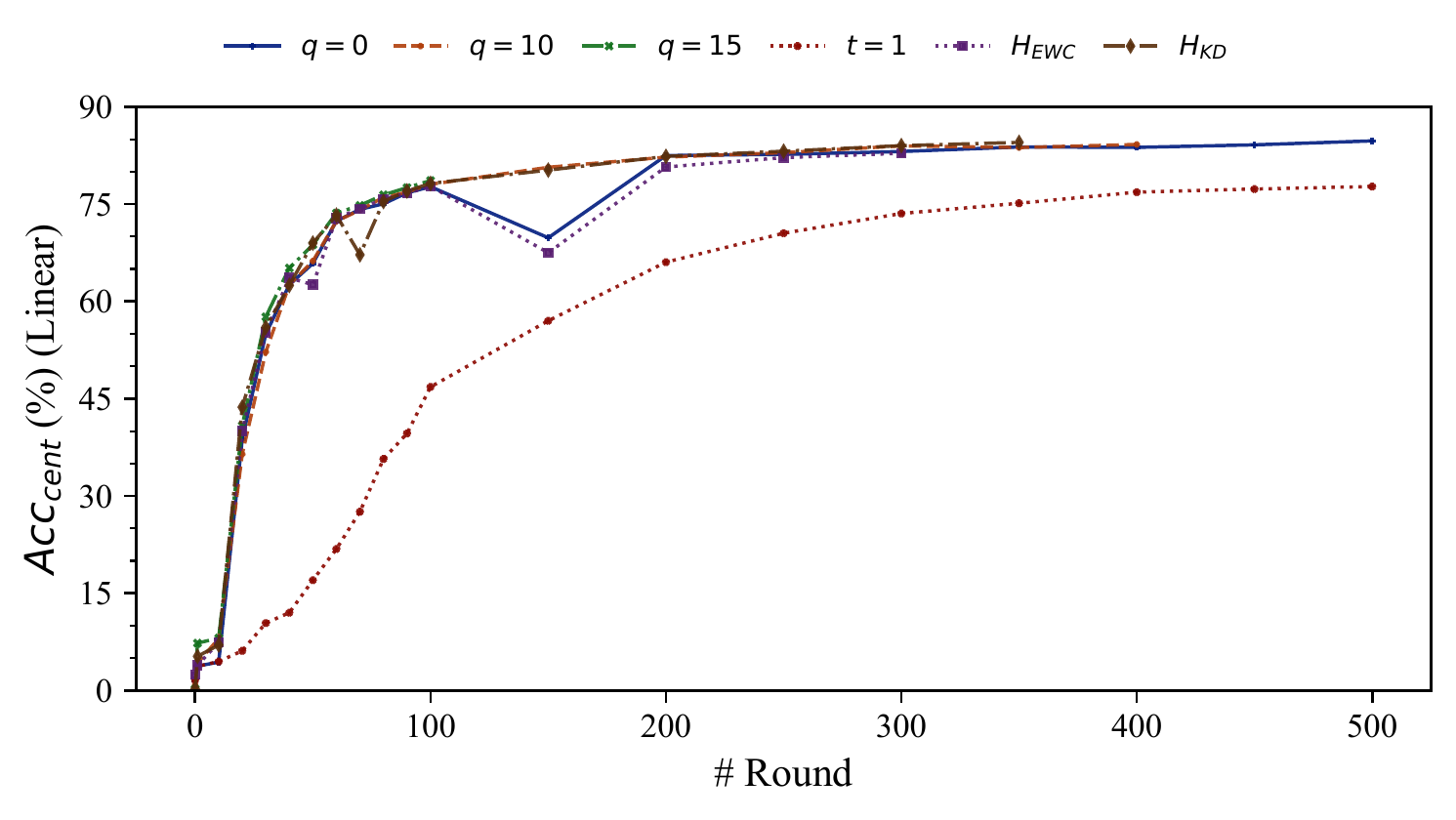}
    \caption{FEMNIST centralised accuracy, q-FFL and PaFl approach FedAvg for our tested hyperparameters while causing increased instability in the training process or outright divergence past a certain round---for such models, we test and adapt the last version before divergence. TERM merely underperforms without divergence for the $t=1$ value.}
    \label{fig1:globalAGGacc:FairPlot:b}
\end{figure}

We begin by examining the convergence process for next-word prediction on Reddit and summarise our findings on the centralised test-set accuracy and local accuracy. As shown in \cref{fig1:globalAGGacc:FairPlot:a} and \cref{tab:res1:fed_perf}, the impact of q-FFL on accuracy is neutral to negative for our tested $q$-values, while that of TERM is highly negative for all tested $t$. Although fairness reliably reduces the accuracy variance for $q \geq 1$, the performance cost is too high for all values reported in \cref{sec:experiments}. Meanwhile, we found that TERM did not obtain acceptable performance for any (t) we explored and excluded it from future Reddit experiments.

\begin{table}[!t]
    \centering
    \resizebox{\columnwidth}{!}{%
        \begin{tabular}{@{}llllllll@{}}
            \toprule
            Objective & $Acc_{cent} (\%)$ & $Avg_{loc} (\%)$ & $B_{loc} (\%)$  & $W_{loc}  (\%)$ & $(Var_{Avg})$  & $(Var_{B})$     & $(Var_{W})$    \\ \midrule
            $q=0$     & 17.826            & 18.645           & 24.572          & 14.815          & 9.177          & \textbf{22.114} & 1.072          \\
            $q=0.1$   & 17.789            & 18.66            & 24.843          & 14.728          & 9.81           & 22.914          & 1.036          \\
            $q=5$     & 14.056            & 14.819           & 20.208          & 11.66           & \textbf{7.983} & 26.769          & 0.69           \\ \midrule
            $t=0.1$   & 14.299            & 16.476           & \textbf{28.985} & 11.806          & 39.584         & 176.42          & \textbf{0.369} \\
            $t=5$     & 14.373            & 16.438           & 28.981          & 11.766          & 39.642         & 175.96          & 0.382          \\ \midrule
            $H_{EWC}$ & 17.322            & 18.255           & 24.653          & 14.226          & 10.277         & 23.059          & 1.185          \\
            $H_{KD}$  & \textbf{18.177}   & \textbf{19.179}  & 26.406          & \textbf{14.887} & 12.438         & 25.85           & 1.039          \\ \midrule
            Local     & NaN               & 4.456            & 10.227          & 1.204           & 8.777          & 31.11           & 0.893          \\ \bottomrule
        \end{tabular}
    }
    \caption[Q-FFL absolute Reddit performance]{Results showing the centralised and local accuracy on Reddit. The  $Acc_{cent} (\%)$ value refers to the accuracy of the federated model on the centralised test set. In contrast, $Avg_{loc} (\%)$, $B_{loc} (\%)$, and $W_{loc} (\%)$ refer to the average accuracy of the model on the local test sets of the whole population, the top $10\%$ of clients in terms of local accuracy and the worst $10\%$ respectively. The $(Var_{Avg})$, $(Var_{B})$, and  $(Var_{W})$ values refer to the variance in accuracy seen by the populations above. While fairness does decrease variance at \(q \geq 1.0\), the harm to accuracy is too great compared to $q=0.1$. The proposed $H_{KD}$ model improves accuracy across clients but increases variance for everyone except the worst performers.}
    \label{tab:res1:fed_perf}
\end{table}

Moving on to the FEMNIST training results, \cref{fig1:globalAGGacc:FairPlot:b} and \cref{tab:res1c:fem_fed_perf_image} show that q-FedAvg and PaFL both tend to cause instability for our hyperparameters. However, both q-FedAvg with \(q=10\) and $H_{KD}$ have a comparable centralised and average local accuracy to FedAvg. Unlike Reddit, TERM is well-behaved at a value of $t=1$ and thus included in future experiments.

\begin{table}[]
    \centering
    \resizebox{\columnwidth}{!}{%
        \begin{tabular}{@{}llllllll@{}}
            \toprule
            Objective & $Acc_{cent} (\%)$ & $Avg_{loc} (\%)$ & $B_{loc} (\%)$  & $W_{loc}  (\%)$ & $(Var_{Avg})$    & $(Var_{B})$   & $(Var_{W})$     \\ \midrule
            $q=0$     & \textbf{84.739}   & 75.341           & 99.435          & 35.717          & 432.507          & 1.151         & 73.747          \\
            $q=10$    & 84.19             & \textbf{76.591}  & 99.013          & \textbf{42.055} & \textbf{320.385} & 1.714         & 64.049          \\
            $q=15$    & 78.634            & 69.749           & 96.681          & 34.988          & 374.637          & 5.177         & \textbf{47.761} \\
            $t=1$     & 77.706            & 69.134           & 98.628          & 33.478          & 417.448          & 2.771         & 56.543          \\ \midrule
            $H_{EWC}$ & 82.825            & 73.964           & 99.321          & 33.457          & 465.605          & 1.376         & 71.481          \\
            $H_{KD}$  & 84.51             & 75.243           & \textbf{99.491} & 34.34           & 443.015          & \textbf{1.07} & 62.91           \\ \midrule
            Local     & NaN               & 46.322           & 92.848          & 0.0             & 1006.77          & 18.144        & 0.0             \\ \bottomrule
        \end{tabular}
    }
    \caption[]{Results for FEMNIST. Unlike Reddit, there did not seem to be a clear proportional relation between accuracy and fairness level for our tested parameters; as such, we chose to report the best and ``fairest'' value. Furthermore, using KD helps the best performers primarily; however, both $H_{KD}$ and $H_{EWC}$ do well.}
    \label{tab:res1c:fem_fed_perf_image}
\end{table}

\cref{tab:res1b:fed_perf_image} indicates the CIFAR-10 image classification task to be more resilient to fairness than previous tasks, with a noticeable accuracy decrease only observable for \(q \geq 10\). The lower sensitivity of this task to FFL is consistent with  \citet{yu2020salvaging} who find CIFAR-10 highly resilient to robust~\citep{yin2018byzantine} FL and differential privacy~\citep{diffPrivate}. Due to the similarity across fairness levels, the convergence graph for CIFAR-10 is not shown. Given this lesser sensitivity for our tested values, we chose not to expand the CIFAR-10 experiments past q-FedAvg and $H_{KD}$. These findings indicate that the dataset heterogeneity may need to be meaningful rather than artificially imposed for significant effects to emerge.

\begin{table}[]
    \centering
    \resizebox{\columnwidth}{!}{%
        \begin{tabular}{@{}llllllll@{}}
            \toprule
            Objective & $Acc_{cent} (\%)$ & $Avg_{loc} (\%)$ & $B_{loc} (\%)$  & $W_{loc}  (\%)$ & $(Var_{Avg})$  & $(Var_{B})$    & $(Var_{W})$    \\ \midrule
            $q=0$     & 81.28             & \textbf{81.37}   & \textbf{82.255} & \textbf{79.864} & 0.568          & 0.022          & 1.067          \\
            $q=5$     & \textbf{81.86}    & 81.221           & 82.011          & 79.794          & \textbf{0.446} & \textbf{0.004} & \textbf{0.643} \\
            $q=15$    & 78.16             & 79.935           & 81.178          & 77.885          & 0.945          & 0.02           & 1.267          \\ \midrule
            Local     & NaN               & 31.718           & 38.297          & 24.649          & 16.3           & 1.543          & 0.906          \\ \bottomrule
        \end{tabular}%
    }
    \caption{ Results for CIFAR-10. Unlike the language task, \(q=5\) represents an optimum across all our tested values in terms of variance while maintaining performance; however, differences are small.}
    \label{tab:res1b:fed_perf_image}
\end{table}

\emph{Implications:} The loss-based averaging mechanism of q-FedAvg and TERM is not guaranteed to improve the final accuracy distribution proportionally to the fairness parameter and may fail to do so under our specific experimental conditions. This calls for further inquiry into the viability of such methods.

\subsection{FFL fails to improve relative accuracy}\label{sec:Results:RelativeImprovement}

\begin{table*}[]
    \begin{tabular}{lllllllll}
        \toprule
        Objective & Adapt     & $Avg_{loc} (\%)$   & \% \textless 0 & $B_{loc}(\%)$      & $W_{loc} (\%)$     & $(Var_{Avg})$      & $(Var_{B})$        & $(Var_{W})$       \\ \midrule
        $q=0$     & $q=0$     & 14.185             & 53             & 20.715             & 9.392              & \textbf{13.323}    & \underline{34.379} & 20.201            \\
                  & A\_FB     & 15.87              & 0              & 25.849             & \textbf{11.311}    & 29.216             & 149.736            & 3.246             \\
                  & A\_EWC    & \textbf{16.046}    & 0              & \underline{27.558} & 11.304             & 36.067             & 178.387            & 3.337             \\
                  & A\_KD     & 15.538             & 0              & 24.376             & 11.209             & 23.112             & 115.016            & \underline{3.183} \\ \midrule
        $q=0.1$   & $q=0.1$   & 14.208             & 50             & 20.907             & 9.359              & \underline{13.742} & \underline{35.005} & 20.733            \\
                  & A\_FB     & 15.827             & 0              & 25.964             & \underline{11.261} & 29.505             & 149.011            & 3.212             \\
                  & A\_EWC    & \underline{15.839} & 0              & \underline{27.692} & 11.024             & 37.108             & 179.066            & 3.336             \\
                  & A\_KD     & 15.546             & 0              & 24.614             & 11.19              & 23.95              & 118.471            & \underline{3.166} \\ \midrule
        $H_{EWC}$ & $H_{EWC}$ & 13.807             & 108            & 20.723             & 8.681              & \underline{14.994} & \underline{38.119} & 24.938            \\
                  & A\_FB     & 15.423             & 0              & 25.709             & \underline{10.88}  & 29.795             & 149.308            & \underline{3.02}  \\
                  & A\_EWC    & \underline{15.561} & 0              & \underline{27.482} & 10.762             & 37.251             & 179.528            & 3.266             \\
                  & A\_KD     & 15.157             & 0              & 24.336             & 10.823             & 23.996             & 117.427            & 3.041             \\ \midrule
        $H_{KD}$  & $H_{KD}$  & 14.729             & 27             & 22.038             & 9.971              & \underline{14.969} & \underline{41.225} & 18.409            \\
                  & A\_FB     & 15.772             & 0              & 26.533             & 11.154             & 31.068             & 150.552            & \textbf{3.156}    \\
                  & A\_EWC    & \underline{15.824} & 2              & \textbf{28.217}    & 10.966             & 38.439             & 178.543            & 3.469             \\
                  & A\_KD     & 15.698             & 0              & 25.358             & \underline{11.214} & 25.367             & 119.418            & 3.241             \\ \bottomrule
    \end{tabular}%
    \caption{Results showing the relative accuracy of FFL and PaFL models on Reddit, $\% < 0$ refers to the number of clients with negative relative accuracy. The best value in a column is bold, while the best in a group is underlined. The chosen optimal fair model does not significantly reduce the number of underperforming clients in our experiments. Alternatively, \(H_{KD}\) lowers it to half. Local adaptation always provides similar results for the chosen hyperparameters.}
    \label{tab1:short_reddit}
\end{table*}

Having established baselines of accuracy for fair models, we can now evaluate the relative accuracy of FFL, PaFL and their interactions with local adaptation. Unfortunately, the CIFAR-10 data is uninformative as the federated model outperforms the local one for all clients, consistent with the findings of \citet{yu2020salvaging}.

\begin{figure*}[]
    \centering
    \begin{subfigure}[t]{1.0\textwidth}
        \centering
        \caption{Reddit relative accuracy, fairness shows no benefit while $H_{KD}$ reduces the number of underperforming clients without adaptation.}
        \includegraphics[width=\textwidth]{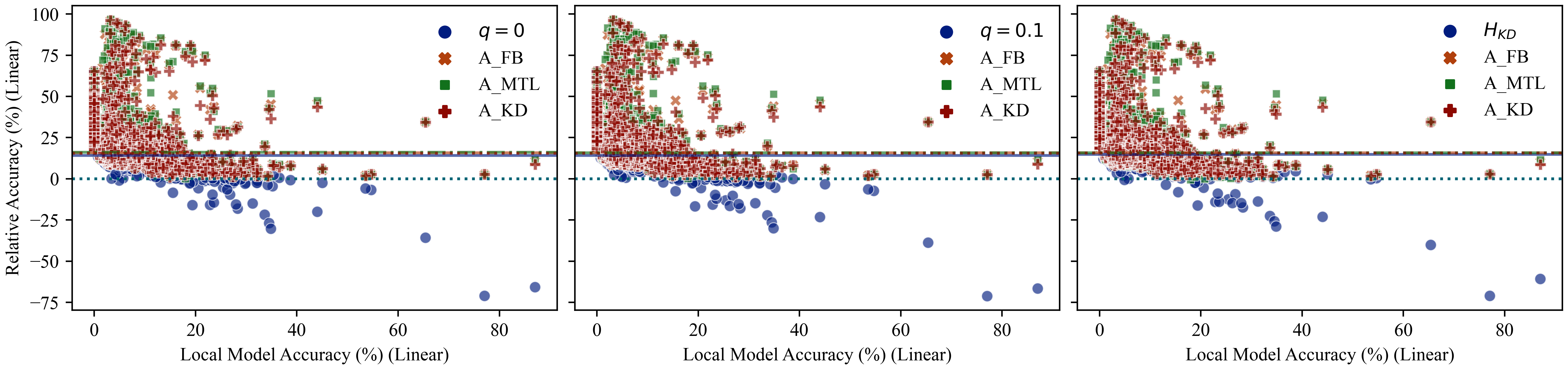}
        \label{fig3:DistribGap:Scatter}
    \end{subfigure}
    \par\bigskip
    \begin{subfigure}[b]{1.0\textwidth}
        \centering
        \caption{FEMNIST results, clients with highly accurate local models are underserved by federated models trained using $q=10$ as many become underperformers. Alternatively, those trained using $H_{EWC}$ receive a slight improvement over FedAvg even when adapted.}
        \includegraphics[width=\textwidth]{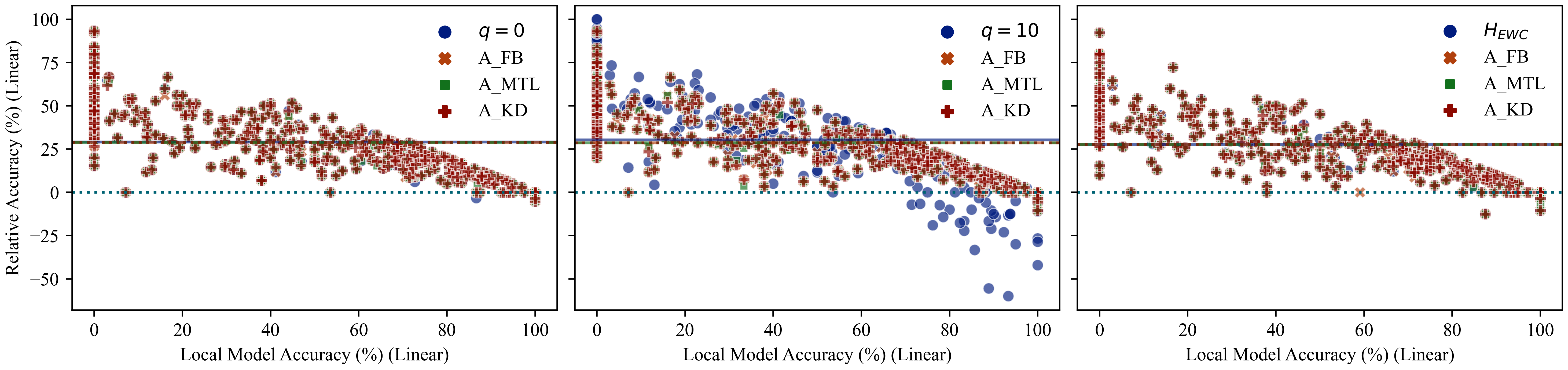}

        \label{fig3:Fem:DistribGap}
    \end{subfigure}
    \caption{Federated or adapted model relative accuracy on a client plotted against local client model accuracy. The horizontal lines represent either the threshold for underperformance ($0$) or the average accuracy of a group of clients.}\label{fig3:DistribGap}
\end{figure*}

For q-FFL, the results for the language task showcased in \cref{tab1:short_reddit} are less than satisfactory as fair models fail to provide benefits in terms of the number of underperforming clients, relative accuracy, or variance. Furthermore, fair models do not offer an improvement over FedAvg once adapted---this is directly visible in the \cref{fig3:DistribGap:Scatter} scatter plot of relative accuracy against local model accuracy.

\begin{table*}[]
    \begin{tabular}{@{}lllllllll@{}}
        \toprule
        Objective & Adapt     & $Avg_{loc} (\%)$   & \% \textless 0 & $B_{loc}(\%)$      & $W_{loc} (\%)$    & $(Var_{Avg})$       & $(Var_{B})$         & $(Var_{W})$         \\ \midrule
        $q=0$     & $q=0$     & \underline{29.02}  & \underline{16} & \underline{65.768} & 2.729             & 338.387             & 137.366             & 12.33               \\
                  & A\_FB     & 28.954             & 17             & 65.463             & 2.72              & \underline{334.754} & \underline{134.824} & 11.853              \\
                  & A\_EWC    & 28.994             & \underline{16} & 65.672             & \textbf{2.802}    & 336.25              & 137.507             & \underline{11.325}  \\
                  & A\_KD     & 28.986             & \underline{16} & 65.684             & 2.788             & 337.14              & 137.796             & 11.594              \\ \midrule
        $q=10$    & $q=10$    & \textbf{30.269}    & 39             & \textbf{79.687}    & -14.844           & 673.351             & \textbf{111.144}    & \underline{206.995} \\
                  & A\_FB     & 28.613             & \textbf{12}    & 64.818             & \underline{2.699} & \underline{320.729} & 123.679             & 15.552              \\
                  & A\_EWC    & 28.612             & 14             & 64.818             & 2.516             & 321.3               & 123.679             & 15.869              \\
                  & A\_KD     & 28.563             & 14             & 64.645             & 2.52              & 320.957             & 127.618             & 15.934              \\ \midrule
        $t=1$     & $t=1$     & \underline{22.812} & 56             & \underline{73.593} & -17.069           & 627.167             & 215.261             & 91.294              \\
                  & A\_FB     & 21.261             & \underline{15} & 50.057             & 0.806             & \textbf{201.35}     & 127.802             & 8.417               \\
                  & A\_EWC    & 21.362             & \underline{15} & 50.218             & 0.765             & 201.703             & \underline{123.192} & 8.102               \\
                  & A\_KD     & 21.202             & \underline{15} & 50.1               & \underline{0.706} & 202.488             & 125.153             & \textbf{7.71}       \\ \midrule
        $H_{EWC}$ & $H_{EWC}$ & \underline{27.642} & \underline{13} & 62.157             & \underline{1.522} & 315.388             & 123.438             & 19.225              \\
                  & A\_FB     & 27.558             & 14             & 62.431             & 1.404             & 316.622             & 124.092             & \underline{18.888}  \\
                  & A\_EWC    & 27.603             & \underline{13} & \underline{62.588} & 1.349             & 315.619             & \underline{122.717} & 20.502              \\
                  & A\_KD     & 27.611             & \underline{13} & 62.588             & 1.542             & \underline{314.112} & \underline{122.717} & 19.176              \\ \midrule
        $H_{KD}$  & $H_{KD}$  & 28.921             & 16             & 66.178             & 1.65              & 353.698             & 154.217             & \underline{16.231}  \\
                  & A\_FB     & 28.916             & 17             & \underline{66.605} & 1.644             & \underline{350.631} & 152.111             & 16.922              \\
                  & A\_EWC    & 28.871             & 16             & \underline{66.605} & 1.719             & 350.742             & 152.111             & 16.865              \\
                  & A\_KD     & \underline{28.967} & \underline{15} & 66.329             & \underline{1.869} & 351.991             & \underline{150.583} & 17.124              \\ \bottomrule
    \end{tabular}%
    \caption{FEMNIST performance of the best fair model and our proposed alternative. Despite providing the highest average relative accuracy and the highest amongst the best \( 10\% \), the model trained using \(q=10\) has more than double the number of underperforming clients of FedAvg ($q=0$). This is also true for TERM with $t=1$. For PaFL, $H_{KD}$ is close to FedAvg while $H_{EWC}$ improves the number of underperforming clients for \emph{both} baseline and adapted models.}
    \label{tab1:shor_fem}
\end{table*}

The results for image recognition on FEMNIST are more unusual yet similarly discouraging for both q-FFL and TERM. \Cref{tab1:shor_fem} makes it clear that the fair model achieves a higher relative accuracy on average and amongst the top \(10\%\) of clients at the cost of obtaining a \emph{negative} relative accuracy on the worst \(10\%\). Additionally, it has over twice as many underperforming clients with negative relative accuracies. We speculate that focusing on clients with high losses harms the accuracy of fair models on those capable of training high-quality local models. This result is corroborated by the final distribution shown in \cref{fig3:Fem:DistribGap}, as \emph{all} the underperforming clients have high local model accuracy. Another factor to consider is the atypical personalisation behaviour of FEMNIST. Models trained with FedAvg and then adapted tend to converge to nearly the same relative accuracy regardless of adaptation technique.

\emph{Implications:} For our tested hyperparameters, Fair FL training algorithms may harm the relative accuracy distribution for datasets where clients can train high-quality local models by themselves. Therefore, we explore a training methodology intended not to sacrifice accuracy for such clients.

\subsection{PaFL as an Alternative}
Having shown the inability of FFL to replace or enhance local adaptation, we argue that it is not the right approach for this application. In principle, for an FL algorithm to provide benefits in terms of relative accuracy, it must achieve two goals. First, it must ensure that the worst-performing clients receive sufficient accuracy to match or exceed local models. Second, for the clients with the \emph{best} local models, it must provide disproportionately high accuracy. While FFL may help fulfil the first requirement, its inability to raise the floor of the worst performers without hurting the ceiling of those that might have an excellent local model makes it incapable of fulfilling the second in our simulations.

Personalisation-aware Federated Learning, in the most general case, offers an alternative where models can be kept closer to one another during training and only allowed to diverge in ways which hurt federated performance the least. Unlike regularisation based on the norm of the distance between model parameters (e.g., FedProx), EWC and KD offer the distinct advantage of determining how a parameter may diverge based on its importance to federated performance. Thus, the model can learn from highly heterogeneous data and raise its accuracy floor for the worst performers without hurting the accuracy ceiling of the best or even improving it.

Preliminary results for the language task are promising in the case of $H_{KD}$ as \Cref{fig1:globalAGGacc:FairPlot:a} and \cref{tab:res1:fed_perf} indicate that it performs better than FedAvg and FFL models in every metric except variance and best-performer variance. Notably, variance is not increased for the worst performers. On the other hand, while $H_{EWC}$ is not far below the FedAvg baseline, it fails to provide any noticeable improvements. In terms of relative accuracy, \cref{tab1:short_reddit} shows that \( H_{KD}\) halves the number of underperforming clients and provides the best average relative accuracy. However, this higher baseline does not translate to improved relative accuracy for adapted models. Overall, lowering the number of clients which \emph{require} adaptation in order to receive an incentive to participate \(H_{KD}\) successfully reduces the need for personalisation on Reddit. On the other hand, \( H_{EWC}\) seems to double the number of underperforming clients for the fixed chosen $\lambda$, although a different value may change results.

For image recognition on FEMNIST, \(H_{KD}\) and \( H_{EWC}\) are satisfactory in terms of centralised and average accuracy according to \cref{fig1:globalAGGacc:FairPlot:b} and \cref{tab:res1c:fem_fed_perf_image}. On the other hand, relative accuracy results in \cref{tab1:shor_fem} are mixed. While both avoid the doubling in underperforming clients that fair models suffer, locally adapted models starting from \( H_{KD} \) as a baseline do not seem to outperform those adapted from FedAvg. Perhaps surprisingly, given its failure on the language task, $H_{EWC}$ reduced the number of underperforming clients for baseline \emph{and} adapted models despite a lower starting average relative accuracy than q-FedAvg and $H_{KD}$. While more experiments are needed, it indicates potential synergy between PaFL and local adaptation.

\emph{Implications:} Personalisation-aware Federated Learning may successfully improve the relative accuracy distribution across all clients. By constraining divergence from the federated model in a manner meaningful to performance, PaFL may learn from clients with ``harder'' datasets without harming the accuracy of those capable of training high-quality local models.
\section{Conclusion}
This paper set out to incentivise FL participation for clients whose local model outperforms a federated one while lowering the need for costly personalisation. Such a reduction would be relevant for federated networks containing devices with limited capabilities for retraining or little data. Our results indicate that FFL is unlikely to provide the desired properties as it did not reduce the number of underperforming clients on Reddit while doubling it on FEMNIST. We hypothesise that Fair FL harms clients on whom the federated model performs well but could train an excellent local model alone. Personalisation-aware Federated Learning offers an alternative approach, allowing loss functions used for local adaptation to be applied during FL and vary across rounds. After partial convergence, we applied EWC or KD to enable learning from worst-performing data without sacrificing performance on the federated distribution. While our chosen EWC configuration did not significantly improve over FedAvg on Reddit, KD showed promising results by reducing the number of underperforming clients by up to $50\%$. Furthermore, both avoided increasing the number of underperforming clients on FEMNIST while EWC slightly lowered it even for adapted models. Unlike more complex systems, which simultaneously train local and federated models, this approach does not require storing an additional model nor keeping it synchronised to the federated one. For KD, the computational overhead is smaller than a local model as it does not require training two separate networks, thus avoiding one of the two backward passes required by systems employing local models requir. For EWC, the advantage is even more apparent as computation scales only in the number of network parameters and requires no additional forward or backward passes to be performed. Consequently, we recommend using it to incentivise participation even when an explicit final local adaptation stage is not applied. In terms of future work, more extensive simulations and the addition of theoretical analyses would allow for greater insight into the relation between fairness and the loss function used during training and adaptation.

\clearpage
\pagebreak

\bibliographystyle{ACM-Reference-Format}
\bibliography{sample-base}

\clearpage
\pagebreak

\end{document}